\documentclass[a4paper]{article}

\usepackage[english]{babel}
\usepackage[utf8x]{inputenc}
\usepackage[T1]{fontenc}

\usepackage[a4paper,top=3cm,bottom=2cm,left=3cm,right=3cm,marginparwidth=1.75cm]{geometry}

\usepackage{amsmath}
\usepackage{graphicx}
\usepackage[colorinlistoftodos]{todonotes}
\usepackage[colorlinks=true, allcolors=blue]{hyperref}

\usepackage{hyperref}
\usepackage{amsfonts} 					
\usepackage{amssymb}					
\usepackage{amsthm}					
\usepackage{newlfont}					
\usepackage{layouts}					
\usepackage{graphicx}					
\usepackage{colortbl}					
\usepackage{wasysym}					
\usepackage{color}                    				
\usepackage{algorithm}
\usepackage[noend]{algpseudocode}

\newcommand{\comment}[1]  {  }

\newcommand{\bz}{{\mathbf z}}
\newcommand{\bx}{{\mathbf x}}
\newcommand{\bk}{{\mathbf k}}

\newcommand{\cZ}{{\mathcal Z}}

\title{Gaussian Mixture Generative Adversarial Networks for Diverse Datasets, and the Unsupervised Clustering of Images}
\author{Matan Ben-Yosef and Daphna Weinshall \\School of Computer Science and Engineering \\ The Hebrew University of Jerusalem, Jerusalem 91904, Israel \\ \texttt{\{matan.benyosef,daphna\}@mail.huji.ac.il}}

\date{}

\begin{document}

\maketitle

\begin{abstract}

Generative Adversarial Networks \cite{2014arXiv1406.2661G} (GANs) have been shown to produce realistically looking synthetic images with remarkable success, yet their performance seems less impressive when the training set is highly diverse. In order to provide a better fit to the target data distribution when the dataset includes many different classes, we propose a variant of the basic GAN model, called Gaussian Mixture GAN (GM-GAN), where the probability distribution over the latent space is a mixture of Gaussians. We also propose a supervised variant which is capable of conditional sample synthesis. In order to evaluate the model's performance, we propose a new scoring method which separately takes into account two (typically conflicting) measures - diversity vs. quality of the generated data.  Through a series of empirical experiments, using both synthetic and real-world datasets, we quantitatively show that GM-GANs outperform baselines, both when evaluated using the commonly used Inception Score \cite{2016arXiv160603498S}, and when evaluated using our own alternative scoring method. In addition, we qualitatively demonstrate how the \textit{unsupervised} variant of GM-GAN tends to map latent vectors sampled from different Gaussians in the latent space to samples of different classes in the data space. We show how this phenomenon can be exploited for the task of unsupervised clustering, and provide quantitative evaluation showing the superiority of our method for the unsupervised clustering of image datasets. Finally, we demonstrate a feature which further sets our model apart from other GAN models: the option to control the quality-diversity trade-off by altering, post-training, the probability distribution of the latent space. This allows one to sample higher quality and lower diversity samples, or vice versa, according to one's needs. 
\end{abstract}

\section{Introduction}
\label{chap:introduction}

Generative models have long been an important and active field of research in machine-learning. Such models take as input a training set of data points from an unknown data distribution, and return an estimate of that distribution. By learning to capture the statistical distribution of the training data, this family of models allows one to generate additional data points by sampling from the learned distribution. Well-known families of generative methods include the Na\"ive Bayes model, Hidden Markov models, Deep Belief Networks, Variational Auto-Encoders \cite{2013arXiv1312.6114K} (VAEs) and Generative Adversarial Networks (GANs) \cite{2014arXiv1406.2661G}.

Generative Adversarial Networks include a family of methods for learning generative models where the computational approach is based on game theory. The goal of a GAN is to learn a Generator ($G$) capable of generating samples from the data distribution ($p_\mathcal{X}$), by converting latent vectors from a lower-dimension latent space ($Z$) to samples in a higher-dimension data space ($\mathcal{X}$). Usually, latent vectors are sampled from $Z$ using the uniform or the normal distribution. In order to train $G$, a Discriminator ($D$) is trained to distinguish real training samples from fake samples generated by $G$. Thus $D$ returns a value $D({\bx}) \in [0,1]$ which can be interpreted as the probability that the input sample (${\bx}$) is a real sample from the data distribution. In this configuration, $G$ is trained to obstruct $D$ by generating samples which better resemble the real training samples, while $D$ is continuously trained to tell apart real from fake samples. 

Crucially, $G$ has no direct access to real samples from the training set, as it learns solely through its interaction with $D$. If $G$ is able to perfectly match the real data distribution $p_\mathcal{X}$, then $D$ will be maximally confused, predicting $0.5$ for all input samples. Such a state is known as a \textit{Nash equilibrium}, and has been shown in \cite{2014arXiv1406.2661G} to be the optimal solution for this learning framework. Both $D$ and $G$ are implemented by deep differentiable networks, typically consisting of multiple convolutional and fully-connected layers. They are alternately trained using the Stochastic Gradient Descent algorithm. 

GANs have been extensively used in the domain of computer-vision, where their applications include super resolution from a single image \cite{2016arXiv160904802L}, text-to-image translation \cite{2016arXiv160505396R}, image-to-image translation \cite{2017arXiv170310593Z, 2016arXiv161107004I, 2017arXiv170305192K}, image in-painting \cite{2016arXiv160707539Y} and video completion \cite{2015arXiv151105440M}. Aside from their usages in the computer-vision domain, GANs have been used for other tasks such as semi-supervised learning \cite{2014arXiv1406.5298K, 2015arXiv151106390S}, music generation \cite{2017arXiv170310847Y, 2017arXiv170906298D}, text generation \cite{2016arXiv160905473Y} and speech enhancement \cite{2017arXiv170309452P}.

In the short period of time since their introduction, many different enhancement methods and training variants have been suggested to improve their performance (see brief review below). Despite these efforts, often a large proportion of the generated samples is, arguably, not satisfactorily realistic. In some cases the generated sample does not resemble any of the real samples from the training set, and human observers find it difficult to classify synthetically generated samples to one of the classes which compose the training set (see illustration in Figure~\ref{fig:low_quality_gan_samples}).

\begin{figure}[!htb] 
\centering \begin{tabular} {ccc}
\includegraphics[width=0.3\linewidth]{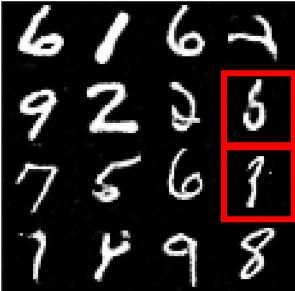} & 
\includegraphics[width=0.3\linewidth]{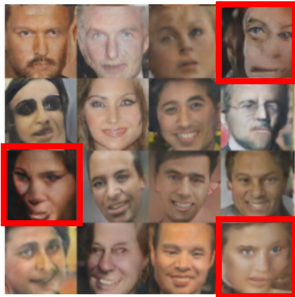} &
\includegraphics[width=0.3\linewidth]{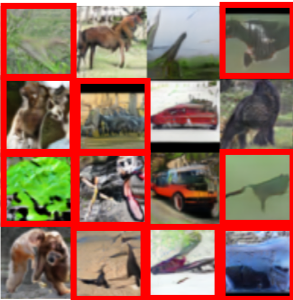} \\
\centering (a) & (b) & (c) \\
\end{tabular}
\caption{Images generated by different GANs trained on (a) MNIST, (b) CelebA, (c) STL-10. Images marked with a red square are, arguably, of low quality.}
\label{fig:low_quality_gan_samples}
\end{figure}
\comment{
\begin{figure}[!htb] 
\centering 
\includegraphics[height=0.4in]{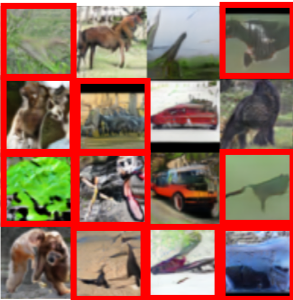} 
\caption{Images generated by different GANs trained on STL-10. Images marked with a red square are, arguably, of low quality.
}
\label{fig:low_quality_gan_samples}
\end{figure}}

The problem described above worsens with the increased complexity of the training set, and specifically when the training set is characterized by large \textit{inter-class} and \textit{intra-class} diversity. In this work we focus on this problem, aiming to improve the performance of GANs when the training dataset has large \textit{inter-class} and \textit{intra-class} diversity.

\comment{\begin{figure}[!htb] 
\centering \begin{tabular} {c}
\includegraphics[width=6.1in]{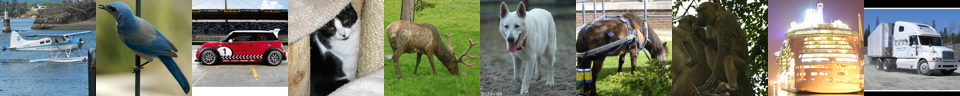} \\ 
(a) \\ \\
\includegraphics[width=6.1in]{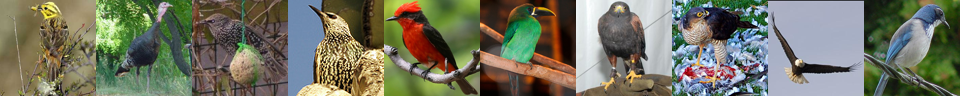} \\
(b) \\
\end{tabular}
\caption{Samples from the STL-10 dataset which demonstrate the complexity of this dataset. (a) Different samples belonging to 10 different classes which demonstrate a large \textit{inter-class} diversity. (b) Different samples belonging to the same class (birds) which demonstrate a large \textit{intra-class}  diversity.}
\label{fig:interclass_intraclass_diversities}
\end{figure}}

\paragraph{Related Work.}
In an attempt to improve the performance of the original GAN model \cite{2014arXiv1406.2661G}, many variants and extensions have been proposed in the past few years.  Much effort was directed at improving GANs through architectural changes to $G$ and $D$, as in the DCGANs described in \cite{2015arXiv151106434R}. Improved performance was reported in \cite{2016arXiv161104076M,2017arXiv170400028G}, among others, by modifying the loss function used to train the GAN model. Additional improvement was achieved by introducing supervision into the training setting, as in conditional GANs \cite{2014arXiv1411.1784M,2016arXiv161009585O}. These conditional variants were shown to enhance the quality of the generated sample, while also improving the stability of the notorious training process of these models.

Another branch of related works, which perhaps more closely relates to our work, involves the learning of a meaningfully structured latent space: Info-GAN \cite{2016arXiv160603657C} decomposes the input noise into an incompressible source and a "latent code", attempting to discover latent factors of variation by maximizing the mutual information between the latent code and the Generator's output. This latent code can be used to discover object classes in a purely unsupervised fashion, although it is not strictly necessary that the latent code be categorical. Adversarial Auto-Encoders \cite{2015arXiv151105644M} employ GANs to perform variational inference by matching the aggregated posterior of the auto-encoder's hidden latent vector with an arbitrary prior distribution. As a result, the decoder of the adversarial auto-encoder learns a deep generative model that maps the imposed prior to the data distribution. \cite{2015arXiv151209300B} combined a Variational Auto-Encoder with a Generative Adversarial Network in order to use the learned feature representations in the GAN's discriminator as basis for the VAE reconstruction objective. As a result, this hybrid model is capable of learning a latent space in which high-level abstract visual features (e.g. wearing glasses) can be modified using simple arithmetic of latent vectors.

\paragraph{Our Approach.}
Although modifications to the structure of the latent space have been investigated before as described above, the significance of the probability distribution used for sampling latent vectors was rarely investigated. A common practice today is to use a standard normal (e.g. $N(0,I)$) or uniform (e.g. $U[0,1]$) probability distribution when sampling latent vectors from the latent space. We wish to challenge this common practice, and  investigate the beneficial effects of modifying the distribution used to sample latent vectors in accordance with properties of the target dataset. 

Specifically, many datasets, especially those of natural images, are quite diverse, with high inter-class and intra-class variability. At the same time, the representations of these datasets usually span high dimensional spaces, which naturally makes them very sparse. Intuitively, this implies that the underlying data distribution, which we try to learn using a GAN, is also sparse, i.e. it mostly consists of low-density areas with relatively few areas of high-density. 

Our approach is to incorporate this prior-knowledge into the model, by sampling latent vectors using a multi-modal probability distribution which better matches these characteristics of the data space. It is important to emphasize that this architectural modification is orthogonal to, and can be used in conjunction with, other architectural improvements such as those reviewed above. Supervision can be incorporated into this model by adding correspondence (not necessarily injective) between labels and mixture components. 

The rest of this paper is organized as follows: In Section~\ref{chap:mm_gan} we describe the family of GM-GAN models. In Section~\ref{chap:gan_evaluation_score} we discuss the shortcomings of the popular Inception Score \cite{2016arXiv160603498S}, and further show that GANs offer a trade-off between sample quality and diversity. We propose an alternative evaluation score which is, arguably, better suited to the task of image synthesis using GANs, and which can quantify the quality-diversity trade-off. In Section~\ref{chap:experimental_evaluation} we empirically evaluate our proposed model in the task of sample synthesis, when trained with various diverse datasets. We show that GM-GANs outperform baselines and achieve better scores. In Section~\ref{chap:unsupervised_clustering} we describe a method for clustering datasets using GM-GANs, and provide qualitative and quantitative evaluation using various datasets of real images.

\section{Gaussian Mixture GAN}
\label{chap:mm_gan}

\paragraph{Unsupervised GM-GAN.}
The target function which we usually optimize for, when training a GAN composed of a Generator $G$ and Discriminator $D$, can be written as follows:
\begin{equation} \label{eq:gan_target_func}
\min_G \max_D V(D,G) = 
	\mathop{\mathbb{E}}_{{\bx} \sim p_\mathcal{X}({\bx})}[\log D({\bx})] + 
    \mathop{\mathbb{E}}_{{\bz} \sim p_{\cZ}({\bz})}[\log (1-D(G({\bz})))]
\end{equation}
Above $p_\mathcal{X}$ denotes the distribution of real training samples, and $p_{\cZ}$ denotes some $d$-dimensional prior distribution which is used as a source of stochasticity for the Generator.
The corresponding loss functions of $G$ and $D$ can be written as follows:
\begin{align} 
\label{eq:g_loss_unsupervised}
L(G) &= -\mathop{\mathbb{E}}_{{\bz} \sim p_{\cZ}({\bz})}[\log D(G({\bz}))] \\
\label{eq:d_loss_unsupervised}
L(D) &=	- \mathop{\mathbb{E}}_{\textbf{x} \sim p_\mathcal{X}(\textbf{x})}[\log D(\textbf{x})]
    - \mathop{\mathbb{E}}_{\textbf{z} \sim p_{Z}(\textbf{z})}[\log (1 - D(G(\textbf{z})))]
\end{align}

Usually, a multivariate uniform distribution (e.g. $U[-1, 1]^d$), or a multivariate normal distribution (e.g. $N(0, I_{d\times d}))$ is used as $p_{\cZ}$ when training GANs.
In our proposed model, we optimize for the same target function as in \ref{eq:gan_target_func}, but instead of using a unimodal random distribution for the prior $p_{\cZ}$, we propose to use a multi-modal distribution which can better suit the inherent multi-modality of the real training data distribution, $p_\mathcal{X}$. In this work, we propose to use a mixture of Gaussians as a multi-modal prior distribution. Formally, we have:
\begin{equation} \label{eq:p_g_gm}
p_{\cZ}({\bz}) = \sum_{k=1}^{K} \alpha_k * p_k({\bz})
\end{equation}
where $K$ denotes the number of Gaussians in the mixture, $\{\alpha_k \}_{k=1}^K$ denotes a categorical random variable, and $p_k({\bz})$ denotes the multivariate Normal distribution $N(\mu_k,\Sigma_k)$, defined by the mean vector $\mu_k$, and the covariance matrix $\Sigma_k$. In the absence of prior knowledge we assume a uniform  mixture of Gaussians, that is, $\forall k\in[K]$ $\alpha_k =\frac{1}{K}$.

The parameters $\mu_k,\Sigma_k$ of each Gaussian in the mixture can be fixed or learned. One may be able to choose these parameters by using prior knowledge, or pick them randomly. Perhaps a more robust solution is to learn the parameters of the Gaussian Mixture along with the parameters of the GAN in an "end-to-end" fashion. This should, intuitively, allow for a more flexible, and perhaps better performing model.  We therefore investigated two variants of the new model - one (static) where the the parameters of the Gaussians mixture are fixed throughout the model's training process, and one (dynamic) where these parameters are allowed to change during the training process in order to potentially converge to a better a solution. These variants are described in detail next:

\begin{description}
\item{\emph{Static GM-GAN.}}
In the basic GM-GAN model, which we call \textit{Static Gaussian Mixture GAN (Static GM-GAN)}, we assume that the parameters of the mixture of Gaussians distribution are fixed before training the model, and cannot change during the training process. More specifically, each of the mean vectors $\mu_k$ is uniformly sampled from the multivariate uniform distribution $U[-c, c]^d$, and each of the covariance matrices $\Sigma_k$ has the form of $\sigma*I_{d \times d}$, where $c\in\mathbb{R}$ and $\sigma \in \mathbb{R}$ are hyper-parameters left to be determined by the user.
\item{\emph{Dynamic GM-GAN.}}
We extend our basic model in order to allow for the dynamic tuning of parameters for each of the Gaussians in the mixture. We start by initializing the mean vectors and covariance matrices as in the static case, but we include them in the set of learnable parameters that are optimized during the GAN's training process. This modification allows the Gaussians' means to wander to new locations, and lets each Gaussian  have a unique covariance matrix. This potentially allows the model to converge to a better local optimum, and achieve better performance. 
\end{description}

The architecture of the \emph{Dynamic} GM-GAN is modified so that $G$ receives as input a categorical random variable $\bk$, which determines from which Gaussian the latent vector should be sampled. This vector is fed into a stochastic node used for sampling latent vectors given the Gaussian's index, i.e. $\textbf{z}|k \sim N(\mu_k, \Sigma_k)$. In order to optimize the parameters of each Gaussian in the training phase, back-propagation would have to be performed through this stochastic node, which is not possible. To overcome this obstacle, we use the re-parameterization trick as suggested by \cite{2013arXiv1312.6114K}: instead of sampling $\textbf{z} \sim N(\mu_k, \Sigma_k)$ we sample $\epsilon \sim N(0,I)$ and define ${\bz}=A_k\epsilon + \mu_k$, where $A \in \mathbb{R}^{d\times d}$ and $\mu_k \in \mathbb{R}^d$ are parameters of the model, and $d$ is the dimension of the latent space. We thus get $\mu({\bz})=\mu_k$ and $\Sigma({\bz})=A_kA_k^T$.

We note that when training either the static or dynamic variants of our model, we optimize for the same loss functions as in (\ref{eq:g_loss_unsupervised}) and (\ref{eq:d_loss_unsupervised}). Clearly other loss functions can be used in conjunction with the suggested architectural modifications, as those changes are independent.

We also note that the dynamic variant of our model includes additional $K * (d^2 + d)$ trainable parameters, as compared to the static model. In cases where $K$ and $d$ are sufficiently large, this can introduce significant computational overhead to the optimization procedure. To mitigate this issue, one can reduce the number of degrees of freedom in $\Sigma_k$, e.g. by assuming a diagonal matrix, in which case the number of additional trainable parameters is reduced to $2*K*d$.

\paragraph{Supervised GM-GAN.}
In the supervised setting, we change the GM-GAN's discriminator so that instead of returning a single scalar, it returns a vector $\textbf{o}\in \mathbb{R}^N$ where $N$ is the number of classes in the dataset. Each element $o_i$ in this vector lies in the range of $[0, 1]$, and can be interpreted as the probability that the given sample is a real sample of class $i$. Informally, this modification can be thought of as having $N$ binary discriminators, where each discriminator $i$ is trained to separate real samples of class $i$ from fake samples of class $i$ and from real samples of classes other than class $i$. 

The Generator's purpose in this setting is, given a latent vector $\bz$ sampled from the $k$'th Gaussian in the mixture, to generate a sample which will be classified by the discriminator as a real sample of class $f(k)$, where $f:[K]\rightarrow[N]$ is a discrete function mapping identity of Gaussians to class labels. When $K=N$, $f$ is bijective and the model is trained to map each Gaussian to a unique class in the data space. When $K>N$ $f$ is surjective, and multiple Gaussians can be mapped to the same class. This can be useful in cases where the training set is characterized by high intra-class diversity and when single classes can be broken down to multiple, visually distinct, sub-classes. When $K<N$ $f$ is injective, and multiple classes can be mapped to the same Gaussian achieving the clustering of class labels. 

We modify both loss functions of $G$ and $D$ to accommodate the class labels. The modified loss functions become the following:
\begin{align*} 
L(G) =& \mathop{-\mathbb{E}}_{\textbf{z} \sim p_{Z}(\textbf{z})} \left[
	\log D(G(\textbf{z}))_{f(y(\bz))} + 
	\sum_{\substack{m=1\\ m\neq f(y(\bz))}}^{N} 		
		\log (1-D(G(\textbf{z}))_m)
	\right] \\
L(D) =&	
    \mathop{-\mathbb{E}}_{\textbf{z} \sim p_{Z}(\textbf{z})} \left[ \sum_{m=1}^{N} \log (1-D(G(\textbf{z}))_m) \right] - 
    \mathop{\mathbb{E}}_{\textbf{x} \sim p_\mathcal{X}(\textbf{x})} \left[
    	\log D(\textbf{x})_{y(\bx)} + 
    	\sum_{\substack{m=1\\ m\neq f(y(\bx))}}^{N}         	
            \log (1 - D(\textbf{x})_m)
    \right]
\end{align*}
where $y({\bx})$ denotes the class label of sample ${\bx}$, and $y({\bz})$ denotes the index of the Gaussian from which the latent vector ${\bz}$ has been sampled. The training procedure for GM-GANs is fully described in Algorithm~\ref{alg:mm_gan_training}. 

\begin{algorithm}[!ht]
	\caption{Training the \textbf{GM-GAN} model.}
	\label{alg:mm_gan_training}
	\begin{algorithmic}[1] 
	
    \Require
		\Statex $K$ - the number of Gaussians in the mixture.
        \Statex $d$ - the dimension of the latent space ($Z$).
        \Statex $c$ - defines the range from which the Gaussians' means are sampled.
        \Statex $\sigma$ - scaling factor for the covariance matrices.
        \Statex $iters$ - the number of training iterations.
        \Statex $b_D$ - the batch size for training the discriminator.
		\Statex $b_G$ - the batch size for training the Generator.        
		\Statex $\gamma$ - the learning-rate.                
        \Statex $f$ - a mapping from Gaussian indices to class indices (in a supervised setting only).
	
    \For{$k=1 ... K$}
		\State Sample $\mu_k \sim U[-c, c]^d$ \Comment init the mean vector of Gaussian $k$
		\State $\Sigma_k \gets \sigma * I_{dxd}$ \Comment init the covariance matrix of Gaussian $k$
	\EndFor
      
	\For{$i=1 ... iters$}
    	\For{$j=1 ... b_D$}
        	\State Sample $\mathbf{x_j} \sim p_{\mathcal{X}}$ \Comment get a real sample from the training-set.                        
            \State Sample $k \sim Categ(\frac{1}{K}, ..., \frac{1}{K})$ \Comment sample a Gaussian index.
            \State Sample $\mathbf{z_j} \sim N(\mu_k, \Sigma_k)$ \Comment sample from the $k$'th Gaussian
            \State $\mathbf{\hat{x}_j} \gets G(\mathbf{z_j})$ \Comment generate a fake sample using the Generator                        
			\If {$supervised$} \Comment compute the loss of $D$
        		\State $L_{real}(D)^{(j)} \gets -\log D(\mathbf{x_j})_{y(\mathbf{x_j})} - 
                \sum_{m=1 , m\neq y(\mathbf{x_j})}^{N} \log (1 - D(\mathbf{x_j})_m)$
                
                \State $L_{fake}(D)^{(j)} \gets -\sum_{m=1}^{N} \log (1-D(\mathbf{\hat{x}_j})_m)$
                
            \Else
            	\State $L_{real}(D)^{(j)} \gets -\log D(\mathbf{x_j})$
            	\State $L_{fake}(D)^{(j)} \gets -\log (1 - D(\mathbf{\hat{x}_j}))$ 
    		\EndIf
    
		\EndFor
        
        \State $L(D) \gets \frac{1}{2*b_D} \sum \limits_{j=1}^{b_D} L_{real}(D)^{(j)} + L_{fake}(D)^{(j)}$
        
        \State $\theta_D \gets Adam(
        	\nabla_{\theta_D}, 
            L(D),
            \theta_D, \gamma
         )$ \Comment update the weights of $D$ by a single GD step.
        
        \For{$j=1 ... b_G$}
        	\State Sample $k \sim Categ(\frac{1}{K}, ..., \frac{1}{K})$ \Comment sample a Gaussian index.
            \State Sample $\mathbf{z_j} \sim N(\mu_k, \Sigma_k)$ \Comment sample from the $k$'th Gaussian
            \State $\mathbf{\hat{x}_j} \gets G(\mathbf{z_j})$ \Comment generate a fake sample using the Generator            
            \If {$supervised$} \Comment compute the loss of $G$   
        		\State $L(G)^{(j)} \gets 
                -\log D(\mathbf{\hat{x}_j})_{f(y(\mathbf{z}_j))} - 
                \sum_{m=1, m\neq f(y(\mathbf{z_j}))}^{N} \log (1-D(\mathbf{\hat{x}_j})_m) $
            \Else
            	\State $L(G)^{(j)} \gets -\log D(\mathbf{\hat{x}_j})$ 
    		\EndIf            
		\EndFor
        
        \State $L(G) \gets \frac{1}{b_G} \sum \limits_{j=1}^{b_G} L_(G)^{(j)}$
        
        \State $\theta_G \gets Adam(
        	\nabla_{\theta_G}, 
            L(G),
            \theta_G, \gamma
         )$ \Comment update the weights of $G$ by a single GD step.
        
	\EndFor
    
    \end{algorithmic}
    
\end{algorithm}

\section{GAN Evaluation Score}
\label{chap:gan_evaluation_score}

We describe next a new scoring method for GANs, which is arguably better suited for the task than the commonly used Inception Score \cite{2016arXiv160603498S}.

\subsection{Inception Score, definition and shortcomings}

\cite{2016arXiv160603498S} proposed a method to evaluate generative models for natural image synthesis, such as VAEs and GANs, using a pre-trained classifier. It is based on the fact that good samples, i.e. images that look like images from the true data distribution, are expected to yield: (i) low entropy $p(y|{\bx})$, implying high prediction confidence; (ii) high entropy $p(y)$, implying highly varied predictions. Here ${\bx}$ denotes an image sampled from the Generator, $p(y|{\bx})$ denotes the inferred class label probability given ${\bx}$ by the Inception network \cite{szegedy2016rethinking} pre-trained on the ImageNet dataset, and $p(y)$ denotes the marginal distribution over all images sampled from the Generator. 

\begin{figure}[ht] 
\begin{tabular} {cc}
\includegraphics[width=0.45\linewidth]{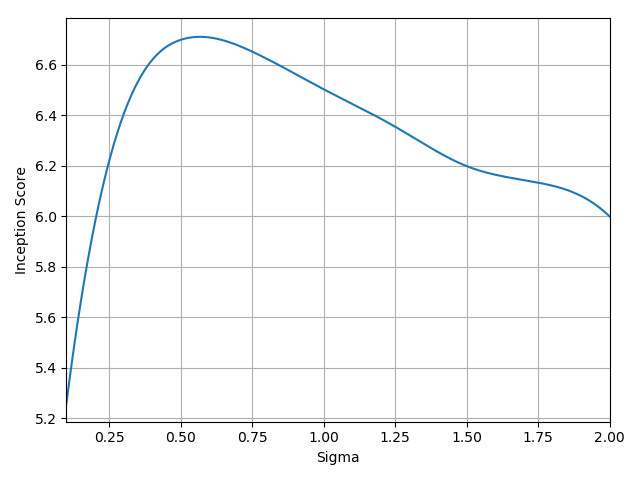} & 
\includegraphics[width=0.45\linewidth]{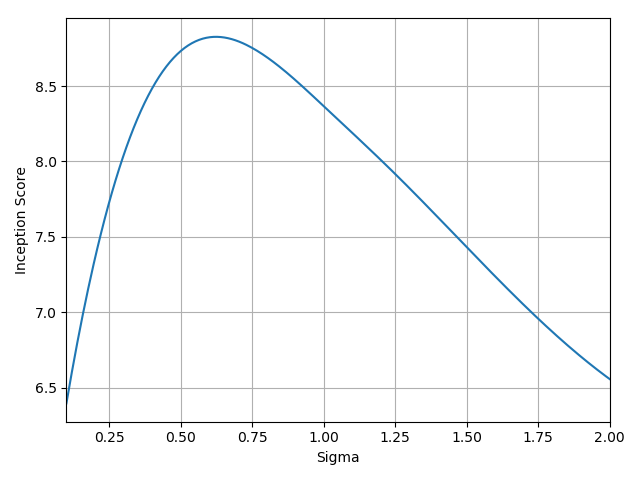} \\
\centering (a) & (b) \\
\end{tabular}
 \caption{Inception Scores of Static GM-GAN models trained on (a) CIFAR-10 and (b) STL-10, when latent vectors are sampled using different values of $\sigma$. In both cases, the same model achieves very different Inception Scores when different values of $\sigma$ are used. Both models were trained using $\sigma=1$. Note that the best score is obtained for $\sigma<1$, far from the training value $\sigma=1$.}
\label{fig:inception_scores_per_sigma}
\end{figure}

The \textit{Inception Score}  \cite{2016arXiv160603498S} is therefore defined as:
\begin{equation}
\exp \left(\mathbb{E}_{\bx \sim p_G} \left[D_{KL}(p(y|\bx) || p(y))\right] \right)
\end{equation}
This score has been used extensively over the last few years. However, it has a number of drawbacks which we found to be rather limiting:
\begin{enumerate}
\item The Inception Score is based on the Inception network \cite{szegedy2016rethinking}, which was pre-trained on the ImageNet dataset. This dataset contains $\sim 1.2$ million natural images belonging to 1,000 different classes. As a result the use of the Inception Score is limited to cases where the dataset consists of natural images. For example, we cannot use the Inception Score to evaluate the performance of a GAN trained on the MNIST dataset, which contains gray-scale images of hand-written digits.
\item Even in cases where the dataset on which we train a GAN consists of natural images, the distribution of these images is likely to be very different from that of ImageNet. In which case, the confidence of the Inception network's prediction on such images may not correlate well with their actual quality.
\item The Inception Score only measures the samples' \textit{inter-class} diversity, namely, the distribution of these samples across different classes $p(y)$. Another equally important measure, which must be taken into account, is the \textit{intra-class} diversity of samples, namely, the variance of different samples which all belong to the same class.
\item The Inception Score combines together a measure of quality and a measure of diversity into a single score. When evaluating the qualities of a GAN using solely this combined score, one cannot asses the true trade-off between the quality and the diversity of generated images. Thus a given Inception Score can be achieved by a GAN which generates very diverse but poor quality images, and also by a GAN which generates similarly looking but high quality images. Different Inception Scores can also be achieved by the same GAN, when sampling latent vectors with different parameters of the source probability distribution (e.g. $\sigma$), as illustrated in Figure \ref{fig:inception_scores_per_sigma}.
\end{enumerate}

\subsection{Alternative Score: Measuring the Quality-Diversity Trade-off}

Our proposed measure is motivated by the expectation that a well trained Generator will map samples of high probability in the latent space to samples of high probability in the target domain, and vice versa. If we measure the quality of a sample ${\bx}\in \mathcal{X}$ in the target domain by its probability $p_\mathcal{X}({\bx})$, then we can expect samples drawn from dense areas in the latent space (i.e. close to the modals of the distribution) to be mapped to high quality samples in the target domain, and vice versa. Therefore, we can increase the expected quality of generated samples in the target domain by sampling with high probability from dense areas of the latent space, and with low probability from sparse areas of the latent space. While increasing the expected quality of generated samples, this procedure also reduces the sample diversity \footnote{In our experiments, we were able to control this quality-diversity trade-off by modifying the probability distribution which is used for sampling latent vectors from the latent space $Z$ (see Figs.~\ref{fig:toy_dataset_quality_diversity_tradeoff}, \ref{fig:mnist_quality_diversity_example}). We further elaborate on this matter in Section~\ref{sec:trade_off_between_quality_and_diversity}.}. This fundamental trade-off between quality and diversity must be quantified if we want to compare the performance different GAN models.

Next we propose a new scoring method for GANs, which allows one to evaluate the trade-off between samples' quality and diversity. This scoring method also relies on a pre-trained classifier, but unlike the Inception Score, this classifier is trained on the \textit{same} training set on which the GAN is trained on. This classifier is used to measure both the quality and the diversity of generated samples, as explained below.

\subsubsection*{Quality Score}
To measure the quality of a generated sample ${\bx}$, we propose to use an intermediate representation of ${\bx}$ in the pre-trained classifier $c$, and to measure the Euclidean distance from this representation to its nearest-neighbor in the training set. More specifically, if $c_l({\bx})$ denotes the activation levels in the pre-trained classifier's layer $l$ given sample ${\bx}$, then the quality score $q({\bx})$ is defined as:

\begin{equation}
\label{eq:quality_score}
q(\bx) = 1 - \frac
	{\exp \left(||c_l({\bx}) - c_l(NN({\bx}))||_2 \right)}
    {\exp \left(||c_l({\bx}) - c_l(NN({\bx}))||_2 \right) + a}
\end{equation}
Above $a$ denotes a constant greater than zero, and $NN({\bx})$ denotes the nearest-neighbor of ${\bx}$ in the training set, defined as $NN({\bx}) = \underset{{\bx}'\in X}{\arg\min} ||c_l({\bx}) - c_l({\bx}')||_2$.
We also define the quality score for a set of samples $X$ as follows: 
\begin{equation}
\label{eq:average_quality_score}
q(X) = \sum_{{\bx}\in X} \frac{1}{|X|} q({\bx})
\end{equation}

\subsubsection*{Diversity Score}
To measure the diversity of generated samples, we take into account both the inter-class, and the intra-class diversity. For \textbf{intra-class} diversity we measure the average (negative) MS-SSIM metric \cite{wang2003multi} between all pairs of generated images in a given set of generated images $X$:
\begin{equation}
\label{eq:intraclass_diversity_score}
d_{intra}(X) = 1 - \frac{1}{|X|^2} \sum_{({\bx}, {\bx}') \in X \times X} MS-SSIM({\bx}, {\bx}')
\end{equation}
For \textbf{intra-class} diversity, we use the pre-trained classifier to classify the set of generated images, such that for each sampled image ${\bx}$, we have a classification prediction in the form of a one-hot vector $c({\bx})$. We then measure the entropy of the average one-hot classification prediction vector to evaluate the diversity between classes in the samples set:
\begin{equation}
\label{eq:interclass_diversity_score}
d_{inter}(X) = \frac{1}{\log(N)} H\left(\frac{1}{|X|} \sum_{{\bx}\in X} c({\bx})\right)
\end{equation}

We combine both the \textbf{intra-class} and the \textbf{inter-class} diversity scores into a single \textbf{diversity score} as follows:

\begin{equation}
\label{eq:combined_diversity_score}
d(X) = \sqrt{d_{intra}(X) * d_{inter}(X)}
\end{equation}

\subsubsection*{Combined Score}

While it is important to look at the \emph{quality} and \emph{diversity} scores separately, since they measure two complementary properties of a model, it is sometimes necessary to obtain a single score per model. We therefore define the following combined measure:
\begin{equation}
\label{eq:combined_score}
s(X) = \sqrt{q(X) * d(X)}
\end{equation}

The range of the proposed \emph{quality}, \emph{diversity} and \emph{combined} scores is $[0, 1]$, where $0$ marks the lowest score, and $1$ marks the highest score. This property makes them easy to comprehend, and convenient to use when comparing the performance of different models.

\section{Experimental Evaluation}
\label{chap:experimental_evaluation}

In this section we empirically evaluate the benefits of our proposed approach, comparing the performance of GM-GAN with alternative baselines. Specifically, we compare the performance of the unsupervised GM-GAN model to that of the originally proposed GAN \cite{2014arXiv1406.2661G}, and the performance of our proposed supervised GM-GAN model to that of AC-GAN \cite{2016arXiv161009585O}. In both cases, the baseline models' latent space probability distribution is standard normal, i.e. $\textbf{z} \sim N(0,I)$. The network architectures and hyper-parameters used for training the GM-GAN models are similar to those used for training the baseline models. For the most part we used the \emph{Static} GM-GAN with default values $d=100$, $c=0.1$, $\sigma=0.15$, $B_D=64$, $b_G=128$, $\gamma=0.0002$; $K$ and $iters$ varied in the different experiments. The \emph{Dynamic} GM-GAN model was only used in Figure~\ref{fig:quality_diversity_scores}.
 
In the following experiments we evaluated the different models on the 6 datasets listed in Table~\ref{table:datasets_descriptions}. In all cases, the only pre-processing made on the training images is a transformation of pixel-values to the range of $[-1,1]$.

\begin{table}[!ht] 
\centering \begin{tabular} {p{2cm}|p{3.5cm}|p{1.5cm}|p{2cm}|p{1.5cm}|p{1.5cm}}

\hline

\textbf{Dataset Name} & \textbf{Description} & \textbf{Number of Classes} &  \textbf{Samples Dimension} &  \textbf{Train Samples} & \textbf{Test Samples} \\
\hline
\hline
  
Toy-Dataset & Points sampled from different Gaussians in the 2-D Euclidean space. & 9 & 2 & 5,000 & - \\
\hline

MNIST  & Images of handwritten digits. & 10 & 28x28x1 & 60,000 & 10,000 \\
\hline

Fashion-MNIST  & Images of clothing articles. & 10 & 28x28x1 & 60,000 & 10,000 \\
\hline

CIFAR-10& Natural images. & 10 & 32x32x3 & 50,000 & 10,000 \\
\hline

STL-10  & Natural images. & 10 & 96x96x3 & 5,000 & 8,000 \\
\hline

Synthetic Traffic Signs & Synthetic images of street traffic signs. & 43 & 40x40x3 & 100,000 & - \\
\hline

\end{tabular}
\caption{Details of the different datasets used in the empirical evaluation: a Toy-Dataset which we have created (see details in Section~\ref{sec:toy-dataset}), MNIST \cite{lecun-mnisthandwrittendigit-2010}, Fashion-MNIST \cite{xiao2017_online}, CIFAR-10 \cite{cifar-10-dataset}, STL-10 \cite{stl-10-dataset} and the Synthetic Traffic Signs Dataset \cite{synthetic-traffic-signs-dataset}. }
\label{table:datasets_descriptions}
\end{table}

\subsection{Toy-Dataset}
\label{sec:toy-dataset}
We first compare the performance of our proposed GM-GAN models to the aforementioned baseline models using a toy dataset, which was created in order to gain more intuition regarding the properties of the GM-GAN model. The dataset consists of 5,000 training samples, where each training sample $\bx$ is a point in $\mathbb{R}^2$ drawn from a homogeneous mixture of $M$ Gaussians, i.e., $\forall \bx$ $p(\bx)=\sum_{m=1}^M \frac{1}{M} p_m(\bx)$ where $p_m(\bx) \sim N(\mu_m,\Sigma_m)$. In our experiments we used $M=9$ Gaussians, $\forall m\in[M]$ $\Sigma_m=0.1*I$ and $\mu=\{-1,0,1\} \times \{-1,0,1\}$. We labeled each sample with the identity of the Gaussian from which it was sampled.

\begin{figure}[ht] 
\begin{tabular} {cc}
\includegraphics[width=2.8in]{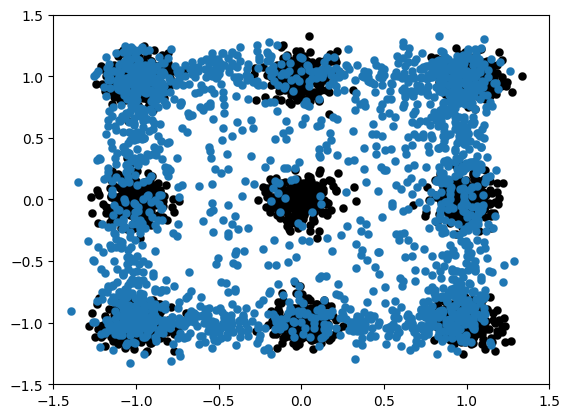} & 
\includegraphics[width=2.8in]{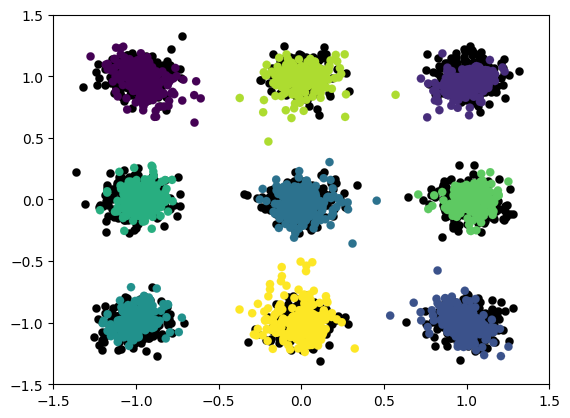} \\
\centering (a) & (b) \\
\includegraphics[width=2.8in]{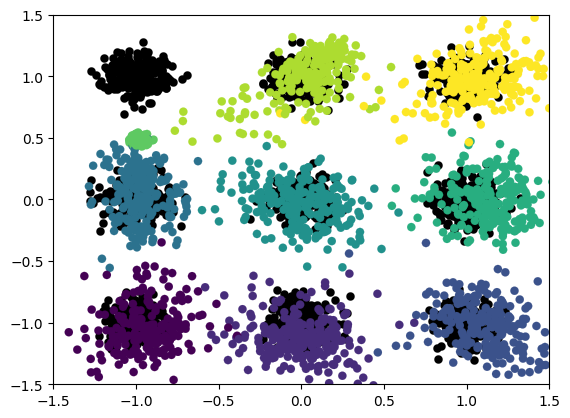} &
\includegraphics[width=2.8in]{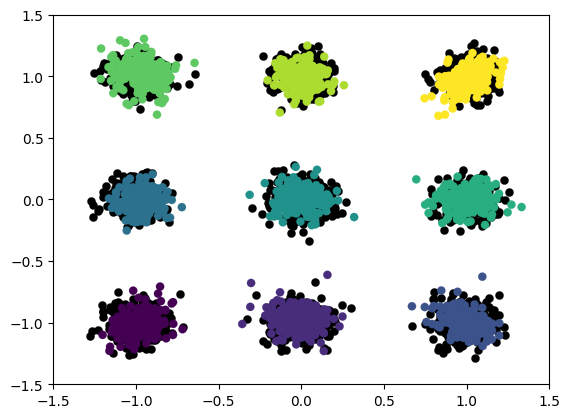} \\
\centering (c) & (d) \\
\end{tabular}
\caption{Samples from the toy-dataset along with samples generated from: (a) GAN, (b) unsupervised GM-GAN, (c) AC-GAN, (d) supervised GM-GAN. Samples from the training set are drawn in black, and samples generated by the trained Generators are drawn in color. In (b) and (d), the color of each sample represents the Gaussian from which the corresponding latent vector was sampled.}
\label{fig:toy_dataset_samples}
\end{figure}

\begin{figure}[htp!] 
\centering \begin{tabular} {ccc}
\centering & GAN & GM-GAN \\

$\sigma=0.25$ & 
\raisebox{-.5\height}{\includegraphics[width=2.0in]{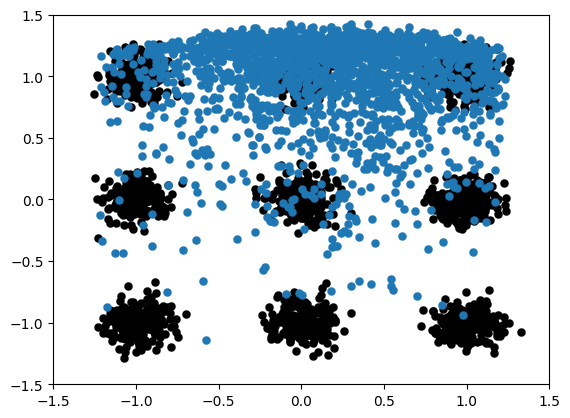}} &
\raisebox{-.5\height}{\includegraphics[width=2.0in]{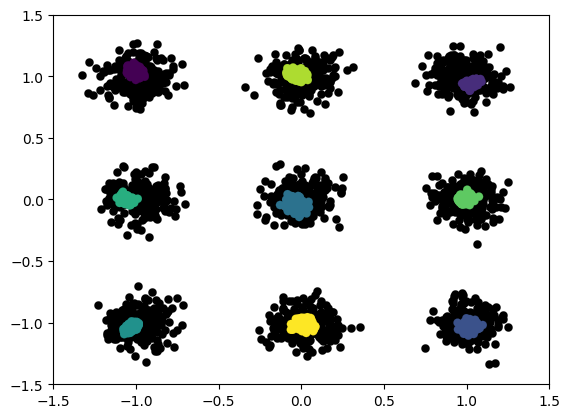}} \\

$\sigma=0.5$ & 
\raisebox{-.5\height}{\includegraphics[width=2.0in]{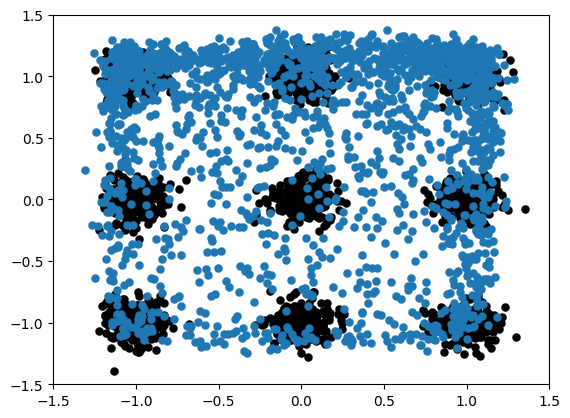}} &
\raisebox{-.5\height}{\includegraphics[width=2.0in]{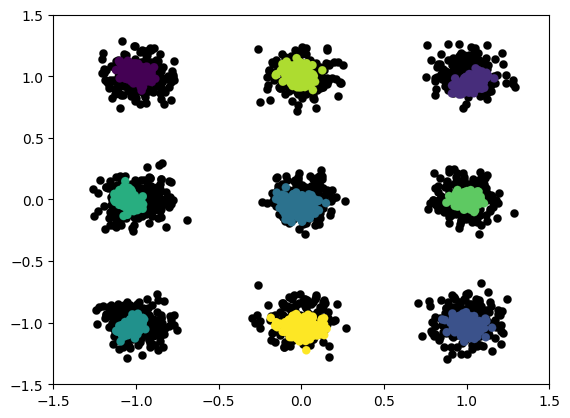}} \\

$\sigma=1.0$ & 
\raisebox{-.5\height}{\includegraphics[width=2.0in]{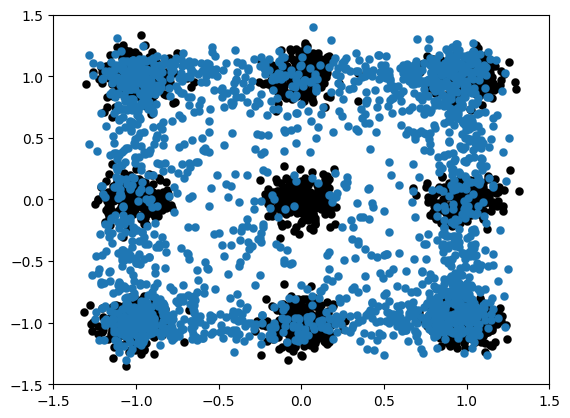}} &
\raisebox{-.5\height}{\includegraphics[width=2.0in]{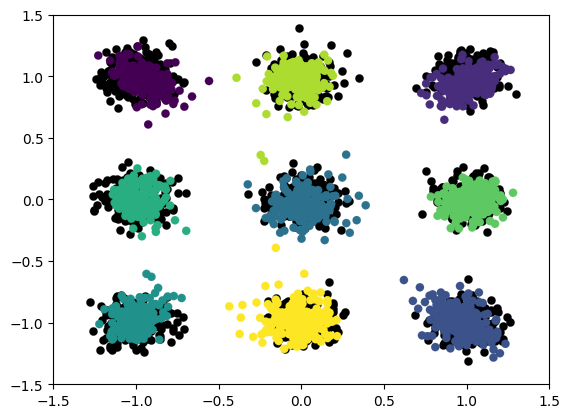}} \\

$\sigma=1.5$ & 
\raisebox{-.5\height}{\includegraphics[width=2.0in]{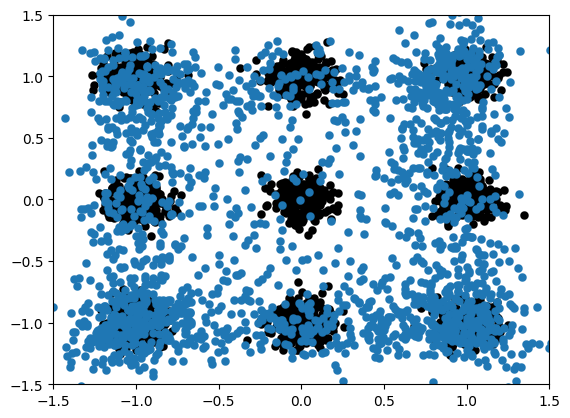}} &
\raisebox{-.5\height}{\includegraphics[width=2.0in]{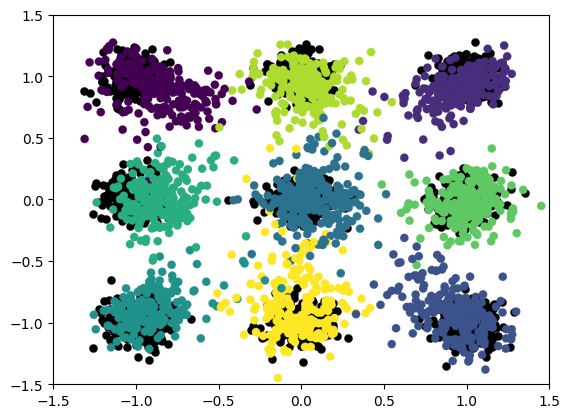}} \\

$\sigma=2.0$ & 
\raisebox{-.5\height}{\includegraphics[width=2.0in]{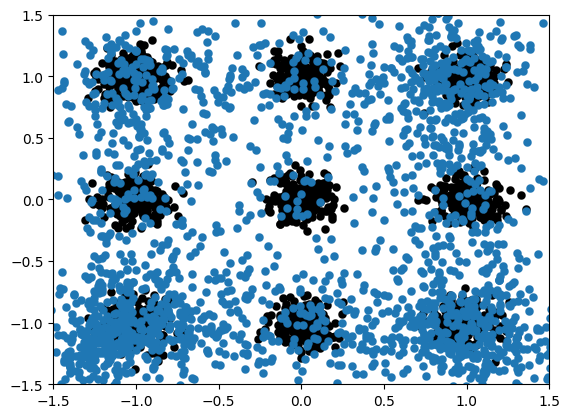}} &
\raisebox{-.5\height}{\includegraphics[width=2.0in]{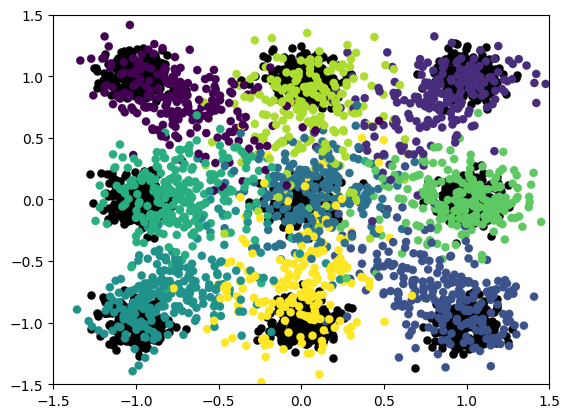}} \\

\end{tabular}
\caption{Samples from the toy-dataset along with samples generated from GAN (left column) and unsupervised GM-GAN (right column), using different $\sigma$ values for sampling latent vectors from the latent space $Z$. During the training process of both models, latent vectors were sampled with $\sigma=1.0$. Samples from the training set are drawn in black, and samples generated by the trained Generators are drawn in color. In samples generated by the GM-GAN, the color of each sample represents the Gaussian from which the corresponding latent vector was sampled. GM-GAN clearly offers a better trade-off between quality and diversity as compared to the baseline.}
\label{fig:toy_dataset_quality_diversity_tradeoff}
\end{figure}

We trained two instances of the GM-GAN model, one supervised using the labels of the samples, and one unsupervised which was not given access to these labels. In both cases, we used $K=9$ Gaussians in the mixture from which latent vectors are sampled. Figure \ref{fig:toy_dataset_samples} presents samples generated by the baseline models (GAN, AC-GAN) and samples generated by our proposed GM-GAN models (both unsupervised and supervised variants). It is clear that both variants of the GM-GAN generate samples with a higher likelihood, which matches the original distribution more closely as compared to the baseline methods. It is also evident that in this configuration, the diversity of samples generated by the GM-GAN model is lower than that of the classic GAN model. This illustrates the trade-off between quality and diversity, which we explore more thoroughly in Section~\ref{sec:trade_off_between_quality_and_diversity}. Figure~\ref{fig:toy_dataset_quality_diversity_tradeoff} demonstrates the superiority of GM-GAN as compared to classic GAN, when measuring the trade-off between quality and diversity offered by these models (see Section~\ref{sec:trade_off_between_quality_and_diversity} for further elaboration on this matter).

\begin{figure}[ht!] 
\centering \includegraphics[width=4.in]{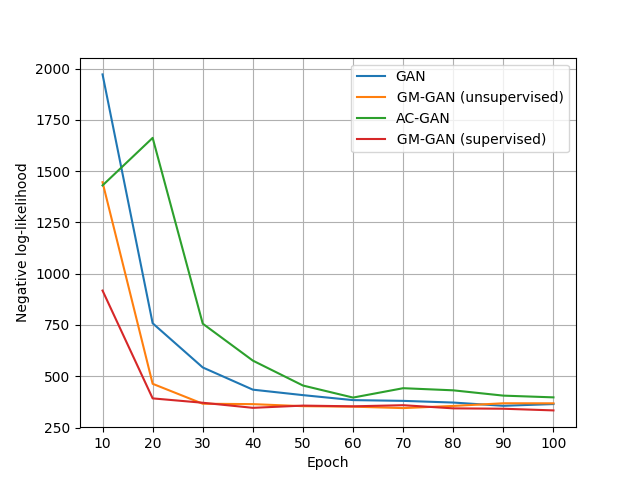}
\caption{Convergence rate of our proposed models vs. baselines. The plot shows the negative log-likelihood of generated samples, as a function of the training epoch of each model. Both variants of the GM-GAN model converge much faster as compared to the baseline models.}
\label{fig:toy_dataset_convergence_rate}
\end{figure}

An intriguing observation is that the GM-GAN's Generator is capable, without any supervision, of mapping each Gaussian in the latent space to samples in the data-space which are almost perfectly aligned with a single Gaussian. We also observe this when training unsupervised GM-GAN on the MNIST and Fashion-MNIST datasets. In Section~\ref{chap:unsupervised_clustering} we exploit this phenomenon by training unsupervised clustering models. Finally, we note that the GM-GAN models converge considerably faster than the classical GAN model. Figure \ref{fig:toy_dataset_convergence_rate} shows the (negative) log-likelihood of samples generated from the different models, as a function of the training epoch.

\subsection{Real Datasets, Inception Scores}
\label{sec:inception_scores}

\begin{table}[!b] 
\centering  \begin{tabular} {cc}

  \textbf{CIFAR-10} & \textbf{STL-10} \\ & \\

  \begin{tabular} {l|l}
    \textbf{Model} (unsupervised) & \textbf{Score} \\
    \hline \hline
    GAN & $5.71$ $(\pm 0.06)$ \\ 
    \hline
    GM-GAN (k=10) & $5.92$ $(\pm 0.07)$ \\
    \hline
    GM-GAN (k=20) & $5.91$ $(\pm 0.05)$ \\
    \hline
    \textbf{GM-GAN (k=30)} & $\textbf{5.98}$ $(\pm \textbf{0.05})$ \\
  \end{tabular} &

  \begin{tabular} {l|l}
    \textbf{Model} (unsupervised) & \textbf{Score} \\
    \hline \hline
    GAN & $6.80$ $(\pm 0.07)$ \\ 
    \hline
    \textbf{GM-GAN (k=10)} & $\textbf{7.06}$ $(\pm \textbf{0.11})$ \\
    \hline
    GM-GAN (k=20) & $6.58$ $(\pm 0.16)$ \\
    \hline
    GM-GAN (k=30) & $7.03$ $(\pm 0.10)$ \\
  \end{tabular}\\ & \\ 

  \begin{tabular} {l|l}
    \textbf{Model} (supervised) & \textbf{Score} \\
    \hline \hline
    AC-GAN & $6.23$ $(\pm 0.07)$ \\ 
    \hline
    \textbf{GM-GAN (k=10)} & $\textbf{6.84}$ $(\pm \textbf{0.03})$ \\
    \hline
    GM-GAN (k=20) & $6.81$ $(\pm 0.04)$ \\
    \hline
    GM-GAN (k=30) & $6.83$ $(\pm 0.02)$ \\
  \end{tabular} &

  \begin{tabular} {l|l}
    \textbf{Model} (supervised) & \textbf{Score} \\
    \hline \hline
    AC-GAN & $7.45$ $(\pm 0.10)$ \\ 
    \hline
    \textbf{GM-GAN (k=10)} & $\textbf{8.32}$ $(\pm \textbf{0.06})$ \\
    \hline
    GM-GAN (k=20) & $8.16$ $(\pm 0.05)$ \\
    \hline
    GM-GAN (k=30) & $8.08$ $(\pm 0.07)$ \\
  \end{tabular}

\end{tabular}
\caption{Inception Scores for different GM-GAN models vs. baselines trained on the CIFAR-10 and STL-10 datasets.}
\label{table:inception_scores}
\end{table}

We next turn to evaluate our proposed models when trained on more complex datasets. We start by using the customary Inception Score \cite{2016arXiv160603498S} to evaluate and compare the performance of the difference models, the two GM-GAN models and the baseline models (GAN and AC-GAN). We trained the models on two real datasets with 10 classes each, the CIFAR-10 \cite{cifar-10-dataset} and STL-10 \cite{stl-10-dataset} datasets. Each variant of the GM-GAN model was trained multiple times, each time using a different number ($K$) of Gaussians in the latent space probability distribution. In addition, each model was trained 10 times using different initial parameter values. We then computed for each model its mean Inception Score and the corresponding standard error. The results for the two unsupervised and two supervised models are presented in Table~\ref{table:inception_scores}. In all cases, the two GM-GAN models achieve higher scores when compared to the respective baseline model. The biggest improvement is achieved in the supervised case, where the supervised variant of the GM-GAN model outperforms AC-GAN by a large margin. We also found that the number of Gaussians used in the GM-GAN's latent space probability distribution can improve or impair the performance of the corresponding model, depending on the dataset. 


\subsection{Trade-off between Quality and Diversity}
\label{sec:trade_off_between_quality_and_diversity}

\begin{figure}[htp!] 
\centering \begin{tabular} {cc}

\includegraphics[width=2.5in]{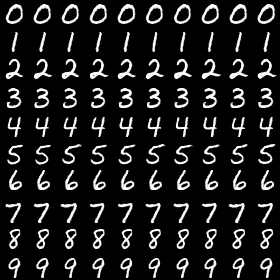} &
\includegraphics[width=2.5in]{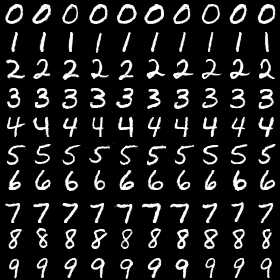} \\
$\sigma=0.1$ & $\sigma=0.4$ \\ \\

\includegraphics[width=2.5in]{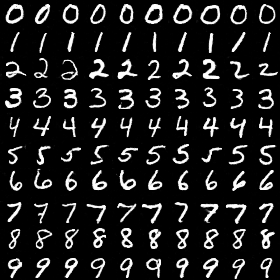} &
\includegraphics[width=2.5in]{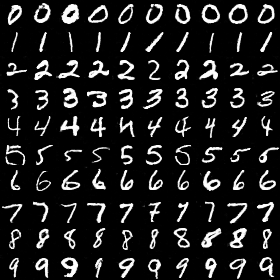} \\
$\sigma=0.7$ & $\sigma=1.0$ \\ \\

\includegraphics[width=2.5in]{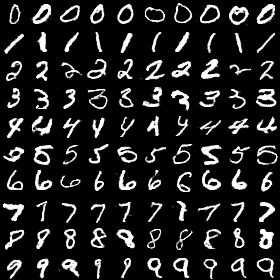} &
\includegraphics[width=2.5in]{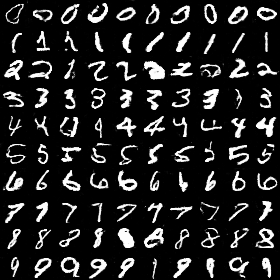} \\
$\sigma=1.5$ & $\sigma=2.0$ \\

\end{tabular}
\caption{Samples taken from an GM-GAN trained on the MNIST dataset. In each panel, samples are taken with a different value of $\sigma$. The quality of samples decreases, and the diversity increases, as $\sigma$ grows.}
\label{fig:mnist_quality_diversity_example}
\end{figure}

As discussed in Section \ref{chap:gan_evaluation_score}, the Inception Score is not sufficient, on its own, to illustrate the trade-off between the quality and the diversity of samples which a certain GAN is capable of generating. In our experiments, we control the quality-diversity trade-off by varying, after the model's training, the probability distribution which is used to sample latent vectors from the latent space. We do so by multiplying the covariance matrix of each Gaussian by a scaling factor $\sigma$. Specifically, when using the baseline models we sample $\textbf{z} \sim N(0, \sigma * I)$, and when using the GM-GAN models we sample $\textbf{z}|k \sim N(\mu_k, \sigma * \Sigma_k)$, $k \sim Categ(\frac{1}{K}, ..., \frac{1}{K})$. Thus, when $\sigma < 1$, latent vectors are sampled with lower variance around the modes of the latent space probability distribution, and therefore the respective samples generated by the Generator are of higher expected quality, but lower expected diversity. The opposite happens when $\sigma > 1$, where the respective samples generated by the Generator are of lower expected quality, but higher expected diversity. Figures \ref{fig:toy_dataset_quality_diversity_tradeoff}, \ref{fig:mnist_quality_diversity_example} demonstrate qualitatively the quality-diversity trade-off offered by GM-GANs when trained on the Toy and MNIST datasets.

\begin{figure}[htp!] 
\centering \begin{tabular} {cc}
\includegraphics[width=0.5\linewidth]{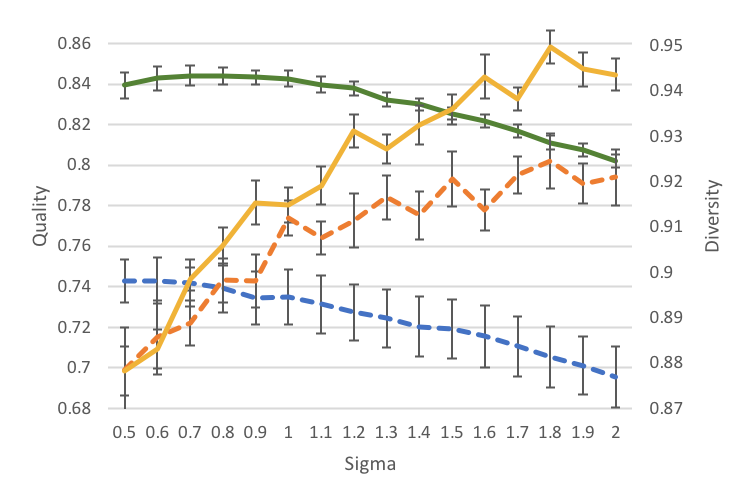} &
\includegraphics[width=0.5\linewidth]{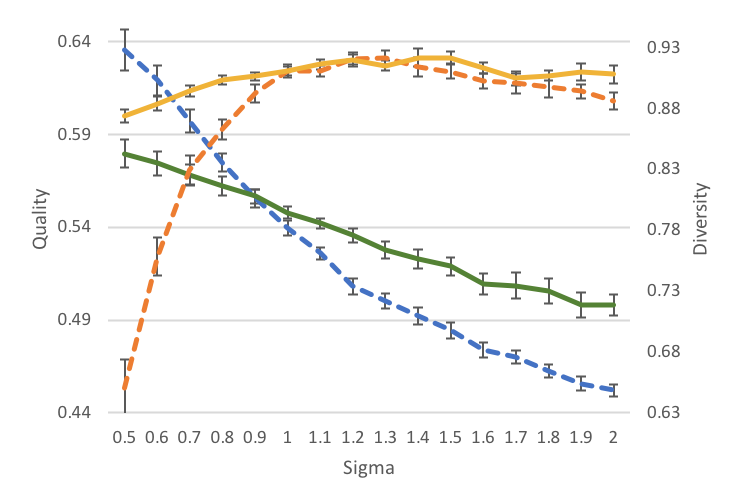} \\
\includegraphics[width=0.5\linewidth]{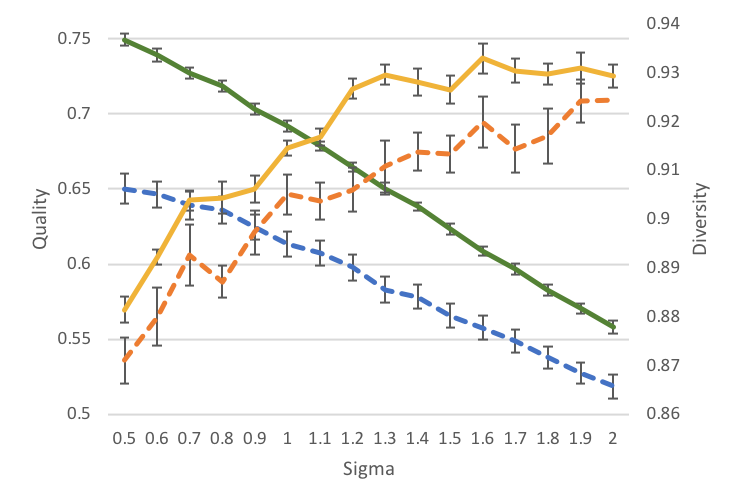} &
\includegraphics[width=0.5\linewidth]{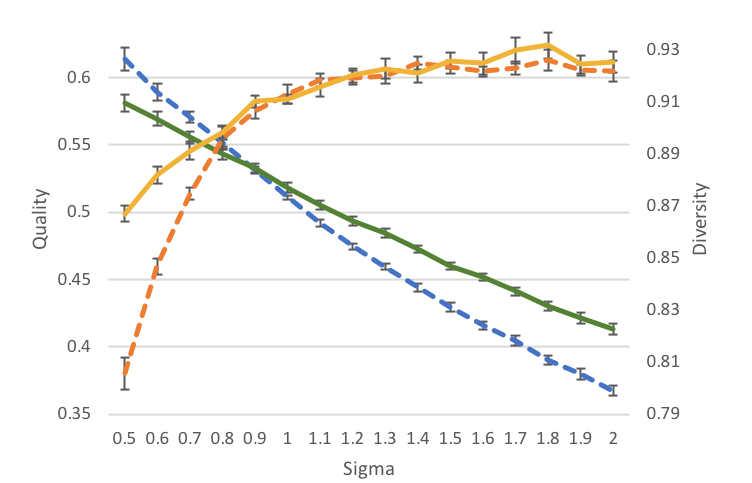} \\
\includegraphics[width=0.5\linewidth]{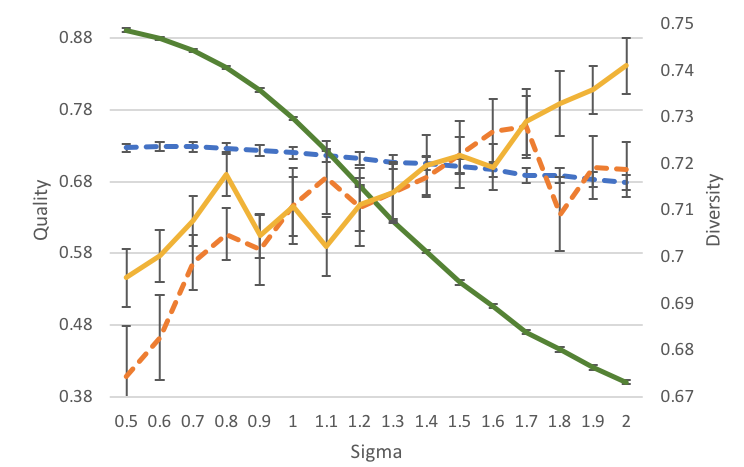} &
\includegraphics[width=0.5\linewidth]{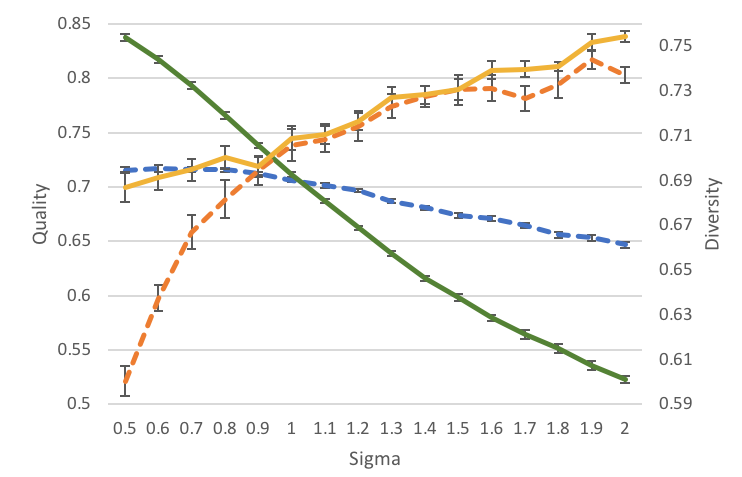} \\
\includegraphics[width=0.5\linewidth]{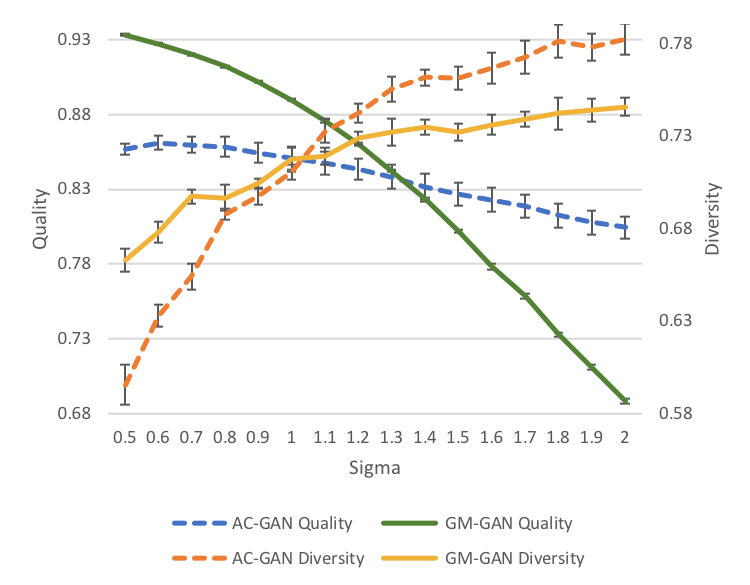} &
\includegraphics[width=0.5\linewidth]{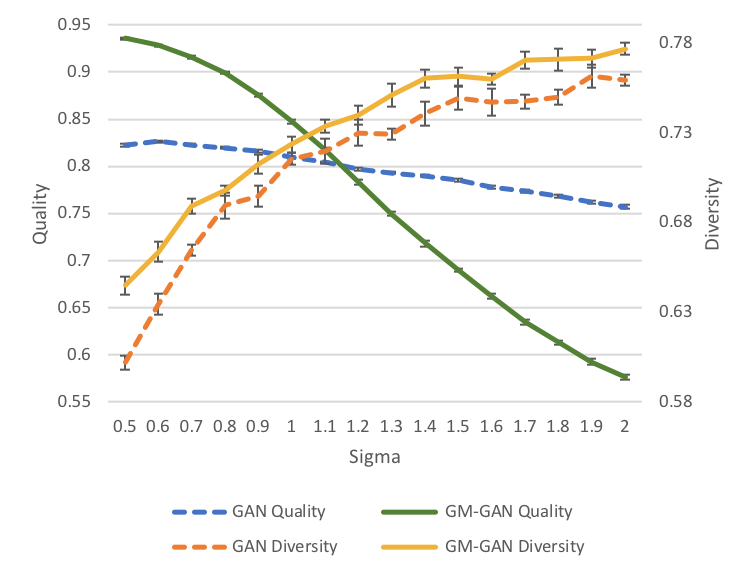}
\end{tabular}
\caption{Quality and Diversity scores of GM-GANs vs. baselines trained on 4 datasets, each corresponding to a different row, shown from top to bottom as follows: \textbf{CIFAR-10}, \textbf{STL-10}, \textbf{Fashion-MNIST} and \textbf{MNIST-10}. Left column: AC-GANs vs. supervised GM-GANs. Right column: GANs vs. unsupervised GM-GANs. Error bars show the standard error of the mean.}
\label{fig:quality_diversity_scores}
\end{figure}

We evaluated each model by calculating our proposed Quality Score from Eq.~(\ref{eq:quality_score}), and the Combined Diversity Score from Eq.~(\ref{eq:combined_diversity_score}), for each $\sigma \in \{0.5, 0.6, ..., 1.9, 2.0\}$. Each model was trained 10 times using different initial parameter values. We computed for each model its mean Quality and mean Combined Diversity scores and the corresponding standard errors. The Quality and Diversity Scores of the GM-GAN and baseline models, when trained on the CIFAR-10, STL-10, Fashion-MNIST and MNIST datasets, are presented in Figure~\ref{fig:quality_diversity_scores}.
In some cases (e.g. supervised training on CIFAR-10 and STL-10) the results show a clear advantage for our proposed model as compared to the baseline, as both the quality and the diversity scores of GM-GAN surpass those of AC-GAN, for \textit{all} values of $\sigma$. In other cases (e.g. unsupervised training on CIFAR-10 and STL-10), the results show that for the lower-end range of $\sigma$, the baseline model offers higher quality, but dramatically lower diversity samples, as compared to our proposed model. In accordance, when visually examining the samples generated by the two models, we notice that most samples generated by the baseline model belong to a single class, while samples generated by our model are much more diverse and are scattered uniformly between different classes. In all cases, the charts predictably show an ascending Quality Score, and a descending Combined Diversity Score, as $\sigma$ is increased.

\section{Unsupervised Clustering using GM-GANs}
\label{chap:unsupervised_clustering}

\begin{figure}[htp!] 
\centering \begin{tabular} {ccc}
\includegraphics[width=0.45\linewidth]{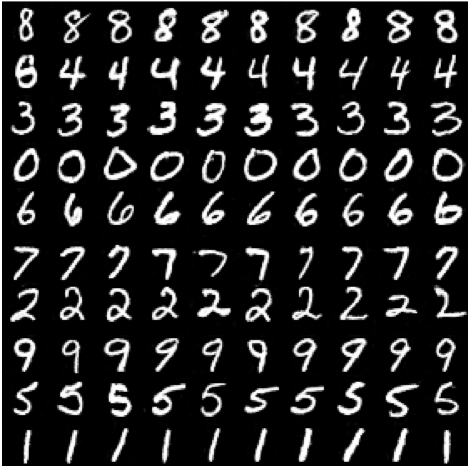} & \hspace{0.01\linewidth} &
\includegraphics[width=0.45\linewidth]{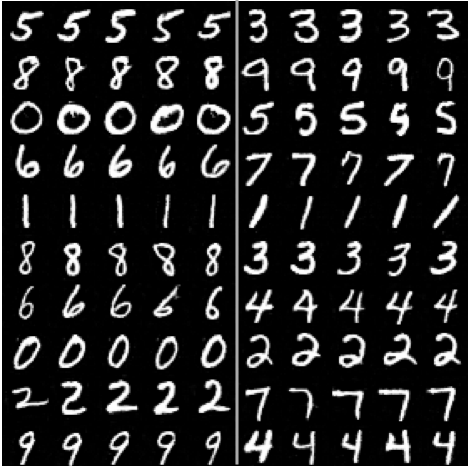} \\
\vspace{0.3cm}

\includegraphics[width=0.45\linewidth]{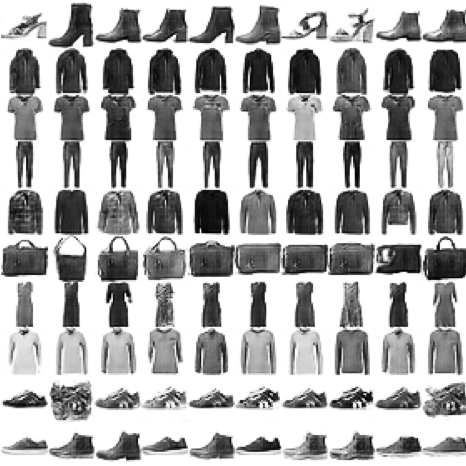} & \hspace{0.01\linewidth} &
\includegraphics[width=0.45\linewidth]{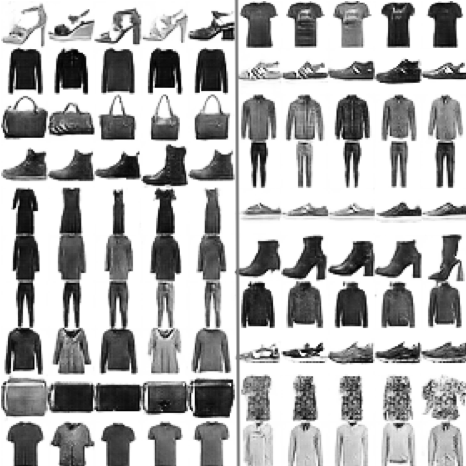} \\
\vspace{0.3cm}

\includegraphics[width=0.45\linewidth]{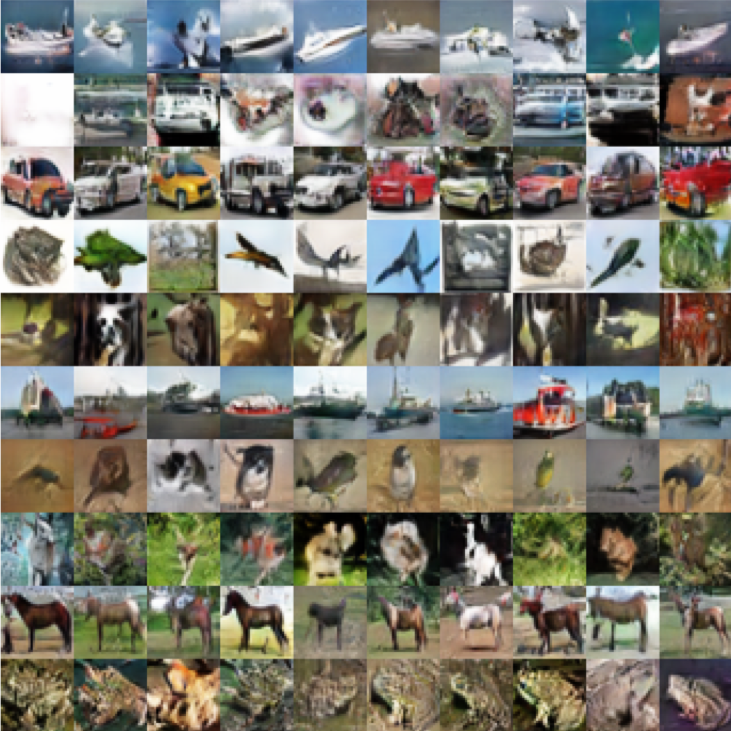} & \hspace{0.01\linewidth} & 
\includegraphics[width=0.45\linewidth]{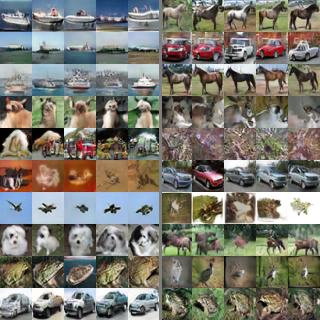} \\
(a) & \hspace{0.01\linewidth} & (b) 
\end{tabular}
\caption{Samples taken from two unsupervised GM-GAN models trained on the MNIST (top panels), Fashion-MNIST (middle panels) and CIFAR-10 (bottom panels) datasets. In (a) the Gaussian mixture contains $K=10$ Gaussians; in each panel, each row contains images sampled from a different Gaussian. In (b) the Gaussian mixture contains $K=20$ Gaussians; in each panel, each half row contains images sampled from a different Gaussian.}
\label{fig:mnist_fashion_mnist_cifar10_unsupervised_mm_gan}
\end{figure}

Throughout our experiments, we noticed an intriguing phenomenon where the \textit{unsupervised} variant of GM-GAN tends to map latent vectors sampled from different Gaussians in the latent space to samples of different classes in the data space. Specifically, each Gaussian in the latent space is usually mapped, by the GM-GAN's Generator, to a single class in the data space. Figures \ref{fig:toy_dataset_samples}, \ref{fig:mnist_fashion_mnist_cifar10_unsupervised_mm_gan} demonstrate this phenomenon on different datasets. The fact that the latent space in our proposed model is sparse, while being composed of multiple Gaussians with little overlap, may be the underlying reason for this phenomenon. Next we exploit this observation to develop a new clustering algorithm, and provide quantitative evaluation of the proposed method.

\subsection{Clustering Method}
\label{sec:clustering_method}
Our clustering method works as follows: we first train an unsupervised GM-GAN on the dataset, where $K$, the number of Gaussians forming the latent space, is set to equal the number of clusters in the intended partition. Using the trained GM-GAN model, we sample from each Gaussian $k \in [K]$ a set of $M$ latent vectors, from which we generate a set of $M$ synthetic samples $\widetilde{X}_k = \left\{\widetilde{\bx}_k^{(i)}\right\}_{i \in [M]}$. We then train a $K$-way multi-class classifier on the unified set of samples from all Gaussians $\bigcup_{k\in [K]} \widetilde{X}_k$, where the label of sample $\widetilde{\bx} \in \widetilde{X}_k$ is set to $k$, i.e. the index of the Gaussian from which the corresponding latent vector has been sampled. Finally, we obtain the soft-assignment to clusters of each sample $\bx$ in the original dataset by using the output of this classifier $c(\bx)\in [0,1]^K$ when given $\bx$ as input. Each element $c(\bx)_k$ ($k \in [K]$) of this output vector marks the association level of the sample $\bx$ to the cluster $k$. A hard-assignment to clusters can be trivially calculated from the soft-assignment vector by selecting the cluster $k$ with which the sample is mostly associated, i.e. $\arg\max_{k\in[K]}c(\bx)_k$. This clustering procedure is formally described in Algorithm \ref{alg:mm_gan_clustering}. 

\begin{algorithm}[ht]
	\caption{Unsupervised clustering procedure using GM-GANs.}
	\label{alg:mm_gan_clustering}
	\begin{algorithmic}[1]
	
    \Require
    	\Statex $X$ - a set of samples to cluster.
		\Statex $K$ - number of clusters.
		\Statex $M$ - number of samples to draw from each Gaussian.
	
    \State $(G,D) \gets$ GM-GAN$(X, K)$ \Comment Train an unsupervised GM-GAN on $X$ using $K$ Gaussians.       
    
    \For{$k=1 ... K$}
		\State Sample $Z_k \sim N(\mu_k, \Sigma_k)^M$ \Comment Sample $M$ latent vectors from the $k$'th latent Gaussian.
        
        \State $\widetilde{X}_k \gets G(Z_k)$ \Comment Generate $M$ samples using the set of latent vectors $Z_k$.                
        
        \State $\forall \mathbf{\widetilde{x}} \in \widetilde{X}_k$ $y(\mathbf{\widetilde{x}}) \gets k$ \Comment Label every sample by the Gaussian from which it was generated.
        
	\EndFor  
    
    \State $\widetilde{X} \gets \bigcup_k \widetilde{X}_k$ \Comment Unite all samples into the set $\widetilde{X}$.
    
    \State $c \gets$ classifier$(\widetilde{X}, y)$ \Comment Train a classifier on samples $\widetilde{X}$ and labels $y$.
    
    \State $\forall \mathbf{x} \in X$ cluster$(\bx) \gets \arg\max_{k \in [K]} c(\mathbf{x})_k$ \Comment Cluster $X$ using classifier $c$.
    
    \end{algorithmic}
\end{algorithm}

\subsection{Empirical Evaluation}
\label{sec:empirical_evaluatrion}
We evaluated the proposed clustering method on three different datasets: MNIST, Fashion-MNIST, and a subset of the Synthetic Traffic Signs Dataset containing 10 selected classes (see Table~\ref{table:datasets_descriptions}). 
To evaluate clustering performance we adopt two commonly used metrics: \textit{Normalized Mutual Information} (NMI), and \textit{Clustering Accuracy} (ACC). \textit{Clustering accuracy} measures the accuracy of the hard-assignment to clusters, with respect to the best permutation of the dataset's ground-truth labels. \textit{Normalized Mutual Information} measures the mutual information between the ground-truth labels and the predicted labels based on the clustering method. The range of both metrics is $[0,1]$ where a larger value indicates more precise clustering results. Both metrics are formally defined as follows:
\begin{align}
\label{eq:acc}
ACC(c|X,y) &= \max_{\pi \in S_N} \frac{1}{|X|} \Sigma_{\mathbf{x}\in X} {\textbf{1}}_{y(\mathbf{x}) = \pi(c(\mathbf{x}))} \\
\label{eq:nmi}
NMI(c|X,y) &= \frac{1}{|X|} \Sigma_{\mathbf{x}\in X} \frac{I(y(\mathbf{x}),c(\mathbf{x}))}{\sqrt{H(y(\mathbf{x})) \; H(c(\mathbf{x}))}}
\end{align}

Above $X$ denotes the dataset on which clustering is performed, $y(\bx)$ denotes the ground-truth label of sample $\bx$, $c(\bx)$ denotes the cluster assignment of sample $\bx$, $H$ denotes entropy, $I$ denotes mutual-information, and $S_N$ denotes the set of all permutations on $N$ elements (the number of classes in the dataset). The unsupervised clustering scores of our method are presented in Table~\ref{table:clustering_accuracy}.

\begin{table}[ht] 
  \centering \begin{tabular}{p{4cm}|p{5cm}|p{2cm}|p{2cm}}
    \hline \hline
    \textbf{Dataset} & \textbf{Method} & \textbf{ACC}  & \textbf{NMI} \\
    \hline \hline
    
    MNIST & K-Means \cite{xie2016unsupervised} & 0.5349 & 0.500 \\
    & AE + K-Means \cite{xie2016unsupervised} & 0.8184 & - \\
    & DEC \cite{xie2016unsupervised} & 0.8430 & - \\     
    & DCEC \cite{guo2017deep}& 0.8897 & 0.8849 \\     
    & InfoGAN \cite{2016arXiv160603657C} & 0.9500 & - \\    
    & CAE-$l_2$ + K-Means \cite{aytekin2018clustering} & 0.9511 & - \\     
    & CatGAN \cite{springenberg2015unsupervised} & 0.9573 & - \\     
    & DEPICT \cite{dizaji2017deep} & 0.9650 & 0.9170 \\
    & DAC \cite{chang2017deep} & 0.9775 & 0.9351 \\
    & GAR \cite{kilinc2018learning} & 0.9832 & - \\
    & IMSAT \cite{hu2017learning} & 0.9840 & - \\
    & \textbf{GM-GAN (Ours)} & \textbf{0.9924} & \textbf{0.9618} \\ 
    \hline       
    
    Synthetic Traffic Signs & K-Means* & 0.2447 & 0.1977 \\         
    & AE + K-Means* & 0.2932 & 0.2738 \\     
    & \textbf{GM-GAN (Ours)} & \textbf{0.8974} & \textbf{0.9274} \\
    \hline    
    
    Fashion-MNIST & K-Means* & 0.4714 & 0.5115 \\       
    & AE + K-Means* & 0.5353 & 0.5261 \\     
    & \textbf{GM-GAN (Ours)} & \textbf{0.5816} & \textbf{0.5690} \\
    \hline   
    
  \end{tabular}
\caption{Clustering performance of our method on different datasets. Scores are based on clustering accuracy (ACC) and normalized mutual information (NMI). Results of a broad range of recent existing solutions are also presented for comparison. The results of alternative methods are the ones reported by the authors in the original papers. Methods marked with (*) are based on our own implementation, as we didn't find any published scores to compare to.}
\label{table:clustering_accuracy}
\end{table}

When evaluated on the MNIST dataset, our method outperforms other recent alternative methods, and, to the best of our knowledge, achieves state-of-the-art performance. Less impressive performance is achieved on the Fashion-MNIST dataset. The fact that this dataset is characterized by small inter-class diversity may be the underlying reason for this. In such a case, an GM-GAN with merely $K=10$ Gaussians may struggle to model this dataset in such a way where different Gaussians in the latent space are mapped to different classes in the data space. Thus, some Gaussians in the latent space are mapped to multiple classes in the data-space and therefore the resulting performance of our method deteriorates. In such cases improved performance can potentially be achieved by increasing the number of Gaussians forming the latent space; however, in this configuration it would not be possible to quantitatively measure the performance of the resulting dataset partitioning, thus we skip this test.

\section{Summary and Discussion}
\label{chap:summary_and_conclusions}

This work is motivated by the observation that the commonly used GAN architecture may be ill suited to model data in such cases where the training set is characterized by large inter-class and intra-class diversity, a common case with real-world datasets these days. To address this problem we propose a variant of the basic GAN model where the probability distribution over the latent space is a mixture of Gaussians, a multi-modal distribution much like the target data distribution which the GAN is trained to model. Additionally, we propose a supervised variant of this model which is capable of conditional sample synthesis. We note that these modifications can be applied to any GAN model, regardless of the specifics of the loss function and architecture.

In order to compare the different models, we note that the performance of GANs, and perhaps other families of generative models, exhibits a certain trade-off between the quality of their generated samples and the diversity of those samples. Therefore arguably the performance of such models must be evaluated by \textit{separately} measuring the quality and the diversity of the generated samples, unlike common practice. For this purpose we propose a scoring method which separately takes into account these two factors. The proposed score can be modified, based on the application's requirement, by adjusting the proportion of each factor when employing the trained model. 

In our empirical study, using both synthetic and real-world datasets, we quantitatively showed that GM-GANs outperform baselines, both when evaluated using the commonly used Inception Score \cite{2016arXiv160603498S}, and when evaluated using our own alternative scoring method. We also demonstrated how the quality-diversity trade-off offered by our models can be controlled, by altering, post-training, the probability distribution of the latent space. This allows one to sample higher-quality, lower-diversity samples or vice versa, according to one's needs. Finally, we qualitatively demonstrated how the \textit{unsupervised} variant of GM-GAN tends to map latent vectors sampled from different Gaussians in the latent space to samples of different classes in the data space. We further showed how this phenomenon can be exploited for the task of unsupervised clustering, and backed our method with quantitative evaluation.

It is important to emphasize that the architectural modifications we proposed in this work are orthogonal to, and can be used in conjunction with, other architectural improvements suggested in prior art, such as those reviewed in Section~\ref{chap:introduction}. Thus, other variants of GANs can also benefit from adopting the proposed method. For example, one may use a multi-modal prior in conjunction with the popular WGAN-GP model \cite{2017arXiv170400028G} in order to achieve better training stability as well as higher quality sample generation, or the InfoGAN model \cite{2016arXiv160603657C} in order to improve the modeling of multi-modal attributes. 

The GM-GAN model, along with the proposed scoring method, allow one to control the quality-diversity trade-off and directly choose between drawing higher-quality or higher-diversity samples. This can be useful in cases where these factors have an influence on the application for which the GAN is employed. For example, when a GAN is used to boost the performance of a classifier trained in a semi-supervised learning settings, e.g. \cite{2015arXiv151106390S, 2016arXiv160603498S}, both the quality and the diversity of the synthetic samples can influence the performance of the target classifier. Thus one may want to carefully choose the right proportions of these two factors when employing the model. Another example is Curriculum Learning \cite{Bengio:2009:CL:1553374.1553380, 2018arXiv180203796W}, a setting in which training samples are gradually revealed from the easiest to the most difficult. Here one can employ our method in order to initially generate high quality and low diversity samples, which are arguably easier, followed by samples of higher diversity and lower quality.

\newcommand{\etalchar}[1]{$^{#1}$}

\bibliographystyle{alpha}

\end{document}